# Ankh ☥: Optimized Protein Language Model Unlocks General-Purpose Modelling

Ahmed Elnaggar[1,2*], Hazem Essam[2], Wafaa Salah-Eldin[2], Walid Moustafa[2], Mohamed Elkerdawy[2], Charlotte Rochereau[3] and Burkhard Rost[1]

[1*]Technical University of Munich (TUM).
[2]Proteinea, Inc.
[3]Columbia University.

*Corresponding author(s). E-mail(s): ahmed.elnaggar@tum.de;
Contributing authors: hazem@proteinea.com; wafaa@proteinea.com; walid@proteinea.com; elkerdawy@proteinea.com; cr3007@columbia.edu; assistant@rostlab.org;

**Abstract**

As opposed to scaling-up protein language models (PLMs), we seek improving performance via protein-specific optimization. Although the proportionality between the language model size and the richness of its learned representations is validated, we prioritize accessibility and pursue a path of data-efficient, cost-reduced, and knowledge-guided optimization. Through over twenty experiments ranging from masking, architecture, and pre-training data, we derive insights from protein-specific experimentation into building a model that interprets the language of life, optimally. We present Ankh, the first general-purpose PLM trained on Google's TPU-v4 surpassing the state-of-the-art performance with fewer parameters (<10% for pre-training, <7% for inference, and <30% for the embedding dimension). We provide a representative range of structure and function benchmarks where Ankh excels. We further provide a protein variant generation analysis on High-N and One-N input data scales where Ankh succeeds in learning protein evolutionary conservation-mutation trends and introducing functional diversity while retaining key structural-functional characteristics. We dedicate our work to promoting accessibility to research innovation via attainable resources.

## 1 Introduction

The analogy between the syntax-semantics of natural languages and the sequence-function of proteins has revolutionized the way humans investigate the language of life [1–8]. Although this analogy is intrinsically valuable when viewed as a precedent step in history leading to the adaptation of NLP's advances on the domain of proteins (e.g., language models), conclusions from the field of NLP do not translate, fully, to protein language. Not only are NLP's model sizes are pursued, it is even proposed that scaling-up protein language models may be significantly more impactful than scaling-up natural language models [9]. The proportionality between the model size and the richness of its learned representations is rather -falsely- encouraged by observing language models of a massive number of parameters trained on a massive number of steps still undergoing notable learning gradient and hence perceived as under-fitted [3, 9, 10]. As a result, opting for more meaningful protein representations or more accurate modeling has gradually shifted to opting for larger models and accordingly, more computational power - less accessibility. Notably, PLM sizes have jumped from $\sim 10^6$ [4] to $\sim 10^9$ [10] parameters recently. Shedding the light, chronologically, on protein language model state-of-the-art (SOTA), we baseline our size-performance benchmark using ProtTrans's ProtT5-XL-U50, an encoder-decoder transformer pre-trained on UniRef50 database whose number of parameters is 3B for training and 1.5B for inference [3]. The evolution of model performance with respect to its size was then demonstrated via RITA, a family of language models taking a first step towards establishing scaling principles for protein sequence modeling. RITA showcases 4 different models with a performance-proportional increase in size from 85M, to 300M, to 680M, to 1.2B parameters [9]. The same trend was then reinforced by ProGen2, a suite of protein language models that are trained on different sequence datasets and whose number of parameters is scaled up to 6.4B [11]. Finally and up to the publication date of this manuscript, the latest contribution promoting model up-scaling is ESM-2, a poll of general-purpose protein language models that also showcase a performance-proportional increase in size from 650M, to 3B, to 15B parameters [10]. The simplified relation between bigger and seemingly-better PLMs, ignores several aspects, including computational costs, task-agnostic model design, and implementation. This raises the research innovation entry barrier and constrains it to scalability. Although model size is, without a doubt, a high impact attribute in pursuing the aforementioned objectives, it is not the only one. The same direction in up-scaling pre-training datasets has proven to be conditional (i.e., bigger datasets are not necessarily better than smaller datasets of higher quality) [3]. We build upon the same direction arguing that up-scaling language models is, too, conditional (i.e., bigger models are not necessarily better than smaller models of protein knowledge-guided means of optimization). In this work, our main objective is to integrate knowledge-guided optimization in an iterative empirical framework that promotes accessibility to research innovation via attainable resources. We title our work "Ankh" (i.e. an Ancient Egyptian symbol denoting the key of life) in analogy to how our model "unlocks" the language of life as learning superior representations of its "letters", the amino acids. This is expanded into two pieces of evidence in evaluating Ankh in terms of optimization and generality. Firstly, surpassing the performance of the SOTA in a representative range of structure and function benchmarks combined with a generation analysis for protein engineering on High-N (family-based) and One-N (single sequence-based) applications, where N refers to the number of input sequences. Secondly, fulfilling this performance via a poll of optimized attributes that not only include the model design but also its development, training, and deployment software and hardware. We provide two pre-trained models referred to as *Ankh_large* and *Ankh_base*, offering two modes of computation depending on the application demands. For convenience, we refer to our main model, *Ankh_large*, as *Ankh*.

## 2 Results

### 2.1 Utmost Computational Efficiency and Utmost Performance

We promote visualizing the performance-size correlation as a trade-off also encompassing computational power. On average, *Ankh* improved the PLM SOTA performance by 4.8% while *Ankh_base* improved it by 3.4% with <10% & 3% of the training parameters and 30% & 15% of the embedding dimension, for *Ankh* and *Ankh_base* respectively (Fig. 1). Since feature extraction is the basis of any subsequent modeling, we compared the time needed in milliseconds to extract the features of a sequence with increasing length up to 1024 residues (Fig. 1). Although the Ankh models support sequence lengths even beyond the maximum length of the pre-defined relative positional embedding dimension, we chose the upper limit of 1024 to accommodate the maximum length supported by the ESM-1b model. We can see that ESM-2 (15B) takes minimally 2.2x & 2.0x and maximally 11.7x & 7.1x the feature extraction time for *Ankh_base* and *Ankh*, respectively.

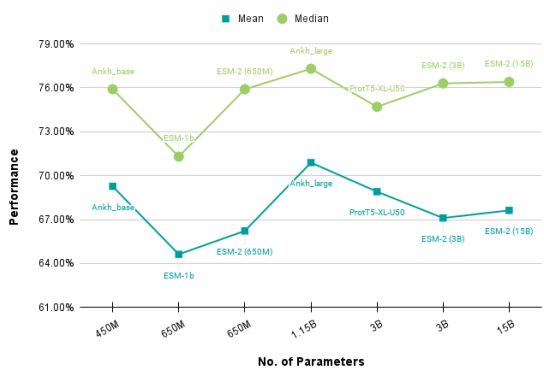

(a)

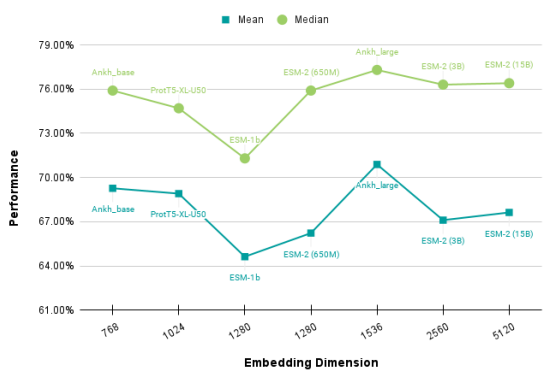

(b)

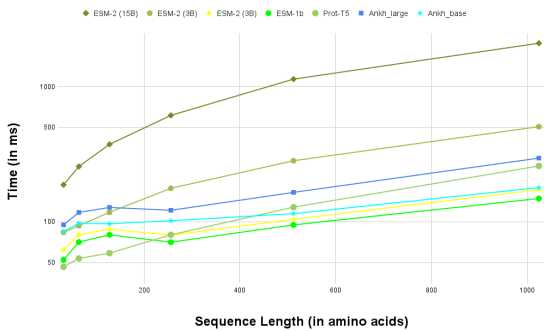

(c)

**Fig. 1**: **Performance-Size Trade-off Comparison:** (a), we plot the number of parameters of different protein language models on the x-axis vs. the mean and median of the performance scores of seven different benchmarking tasks on the y-axis. (b), we plot the embedding dimensions of the investigated models on the x-axis vs. the same y-axis as (a). In diagram (c), we plot the increasing sequence length up to 1024 amino acids on the x-axis and on the y-axis, we plot the corresponding feature extraction time in ms for all of the investigated models.

As for the storage, we needed four A100 80GB Tensor Core GPUs for the feature extraction using ESM-2 (15B) compared to a single A100 40 GB Tensor Core GPU for each of the Ankh models. These computational attributes showcase that besides achieving the top average and median downstream performance, Ankh offers a significantly more accessible computational demand. For the reported results on the protein benchmarking tasks in Table 1, the contextualized embeddings of the protein sequences of each dataset are extracted from the last hidden states of all the investigated models. Our work promotes embedding extraction over attention extraction as means of transfer learning in light of promoting computational optimization. Therefore, we optimized our experimentation with respect to embedding-based predictions. However and as attention maps are reported as the better indicator of contact prediction for the ESM PLM family in [2] and [10], the two representations are tested, separately. To elaborate, the attention maps are extracted and compared with the contextualized embeddings as input for the contact prediction task per every model to opt for fair comparison and demonstrate the SOTA best indicator with what we deem as the best indicator. Indeed, the results show significant out-performance via embedding-based prediction against attention-based prediction (the full results of attention-based predictions can be found in Table 13. As shown in 1, the Ankh suite consistently outperformed the rest of the investigated models. We also observe that ESM-2 (15B) did not outperform the smaller models from the ESM family in all of the tasks in addition to showing inconsistent results across different runs, affirming our hypothesis that bigger models are not necessarily better in all protein modeling tasks and that extensive model sizes bring out its own challenges.

## 2.2 Protein Generation Made Attainable

### 2.2.1 Auto-Regressive Fine-Tuning for High-N Generation

We propose an auto-regressive fine-tuning generation framework for the High-N (protein family-based generation) scale as it offers an accessible

**Table 1**: Results Summary.

| Task | Dataset | Ankh | Ankh_base | ProtT5-XL-U50 | ESM-1b | ESM-2 (650M) | ESM-2 (3B) | ESM-2 (15B) |
|------|---------|------|-----------|---------------|--------|--------------|-------------|--------------|
| **SSP** | *CASP12* [41] | **83.8**±3% | 80.8±4% | 83.4±4% | 79.6±4% | 82.3±4% | 83.3±4% | 83.2±3% |
| | *CASP14* [42] | **77.6**±3% | 76.8±3% | 74.1±3% | 75.1±4% | 77.0±3% | 76.8±3% | 76.8±4% |
| **CP** | *ProteinNet L/1* [34] | **49.0**±8% | 43.2±8% | 44.7±8% | 30.0±6% | 29.6±6% | 30.7±6% | 33.3±6% |
| | *ProteinNet L/5* | **73.2**±11% | 66.6±11% | 69.2±11% | 50.1±10% | 50.2±10% | 52.7±10% | 54.7±10% |
| | *CASP14 L/1* | **30.2**±8% | 28.8±7% | 26.9±7% | 24.6±6% | 25.0±6% | 24.8±7% | 25.9±7% |
| | *CASP14 L/5* | **50.7**±11% | 48.0±11% | 42.4±14% | 40.0±11% | 38.4±13% | 41.9±14% | 40.4±15% |
| **EAT** | | 71.7±6% | **74.8**±6% | 71.0±6% | 64.5±7% | 55.5±7% | 65.6±6% | 65.4±7% |
| **FolP** | | **61.1**±4% | 58.8±4% | 57.6±4% | 57.6±4% | 56.3±4% | 60.5±4% | 56.7±4% |
| **FluP** | | **0.62**±0.004 | 0.61±0.004 | 0.58±0.004 | 0.5±0.005 | 0.48±0.005 | 0.48±0.005 | 0.55±0.004 |
| **SolP** | | **76.4**±2% | 74.2±2% | 74.4±2% | 67.3±2% | 75.0±2% | 74.9±2% | 60.4±2% |
| **GB1P** | | 0.84±0.008 | **0.85**±0.008 | 0.78±0.01 | 0.81±0.009 | 0.82±0.009 | 0.81±0.009 | 0.57±0.02 |
| **LocP** | | **83.2**±2% | 81.4±2% | **83.2**±2% | 80.0±2% | 81.8±2% | 82.4±2% | 81.8±2% |

**Summary of the benchmarking results involving,** *Ankh* **and** *Ankh_base*, **with ProtT5-XL-U50, ESM-1b, ESM-2 (650M), ESM-2 (3B), ESM-2 (15B)**: We report the Spearman Correlation score for the regression tasks and accuracy scores for all classification tasks, except for contact prediction where we report the L/1 and L/5 precision. In EAT, the score reported is the mean of the accuracy scores of the four annotations (Class, Architecture, Topology and Homologous super-family). Task Abbreviations: SSP: Secondary Structure Prediction; CP: Contact Prediction; EAT: Embedding-based Annotation Transfer; FolP: Fold Prediction; FluP: Fluorescence Prediction; SolP: Solubility Prediction; GB1P: GB1 Fitness Prediction ;LocP: Localization Prediction approach for protein variant generation that can be easily scaled across different protein families. Moreover, the framework offers easy control over the generation's exploration-exploitation trade-off by manipulating the logit warping temperature, a parameter used NLP models to increase/decrease the model's confidence in its most likely response [12]. To validate the framework and select the best temperature value range, the same model, fine-tuned on malate dehydrogenase (MDH) natural variants, was utilized to generate three initial sets of 500 sequences with three different temperatures (1.0, 1.5, and 2.0). For the three sets, the Shannon entropy variations between the generated set and a multi-sequence alignment (MSA) of the fine tuning data are reported. Shannon Entropy aims to characterize the generated sequences' preservation of the evolutionary properties of the natural dataset by comparing their representative statistics of amino acid variation. Low entropy values reflect conserved regions governing retained functionality whereas high entropy values reflect less conserved regions with higher mutation rates. In Figure 2 (a), we can observe that the three generated sets show high similarity to the distribution of the natural sequences with almost identical positions for both peaks and valleys. The mean square error (MSE) between entropy values of natural and generated sets are quantified as 0.1, 0.09, and 0.08 for a generation with temperatures 1.0, 1.5, and 2.0 respectively. We emphasize that the reported similarity is calculated based on generated set of 500, which is less than 3% of the total sequences of the natural set (16,706). In other words, the model can mimic the distribution of the fine-tuning dataset with a small portion of its original sequences. To further investigate the effect of different temperatures on the exploration-exploration trade-off, we focus on the generated sequences via temperatures 1.0 and 2.0 due to their direct correlation to introducing functional diversity whilst maintaining conserved functional regions as observed. With this regard, we first report a comparison of the global alignment based identity between the generated sequences and the fine-tuning dataset. We compute the identity via BioPython's pairwise2 module where we obtain the global alignment between the generated variants (gen) and the original ones (nat) [13]. We can observe in Figure 2 (d) that the generated sequences span a wide range of global alignment-based identity scores with sequences as different as 70% and 55% for the generation with temperatures 1.0, and 2.0, respectively. Moreover, over 95% of the generated variants were unique (i.e., only 5% of the variants were duplicates of the fine-tuning dataset sequences). Furthermore, we report the internal identity of each set where we obtain the global alignment between the sequences of each set and themselves. We can observe in Figure 2 (e) that the generated set at 1.0 temperature show less internal variability than the natural sequences, while the generated set at 2.0 shows higher internal variability. Consequently, generation at temperature 1.0 tends to be more conservative, favoring similarity to the most dominating clusters in the fine-tuning dataset. On the other hand, generation at 2.0 tends to be less conservative, covering rare clusters and presenting more diversity in the generation. This allows the user control over the generation process according to the nature of the fine-tuning protein family, and the interest in the generation of global or local variants. Focusing on comparing natural and generated domains with known structural annotations, we observe that the CATH domains in natural variants dominantly belong to three homologous super-families. In all of the generated sets, a significant percentage of the generated sequences includes domains from the three major super-families as shown in Fig. 2 (f) and detailed numerically in Table 15. The domains with known functional classifications are further investigated to compare the functional diversity of the generated sequences. Natural domains from only two of the three homologous super-families (3.90.110.10 and 3.40.50.720) are annotated with functional-family numbers. The functionally annotated CATH domains belonging to the 3.90.110.10 super-family are visualized in Fig 2 (g). All of the generated sets contain domains belonging to the common functional-family numbers: 2, 11, and 3. However, domains belonging to the rare family number 1 (only 6 occurrences in the natural set) can be only observed in sequences generated at temperature 2 (155 and 95 occurrences in epochs 1, and 2 respectively). The same distribution trend is also conserved in the functionally annotated domains belonging to the super-family 3.40.50.720 as can be seen in Table 16

### 2.2.2 Masked Language Modeling (MLM) for One-N Generation

Since we utilize the Masked Language Modeling (MLM) generation framework on the One-Shot (single-sequence generation) scale, it is infeasible to evaluate the generated sequences w.r.t. a specific dataset. Instead, we evaluate the retention of their experimental 3D structural -and accordingly functional- characteristics. To fulfill this, we use ColabFold's $colabfold\_batch$ with 2 models and 3 cycles to predict the 3D structure of the generated variants [14]. Per every generated sequence, we plot its identity to the original unmasked sequences vs. the root mean square deviation (RMSD) between its predicted structure and the experimental 3D structure of the original sequences. We compute the RMSD from the C-alpha atom via BioPython's SVDSuperimposer [13]. Ideally, we would want our model to retain the semantics of the sequence while changing its syntax. In other words, we would want to have variants with low sequence identities as well as low RMSD. To test this assumption, we generate 179 synthetic variants utilizing two Masking Probability 40% and 50% of the input dataset. Indeed, we can see in Figure 3 that the model was utilized to generate many sequences with low sequence similarity while maintaining high structural similarity compared with the input sequences. In the figure, we can notice that sequences generated by 50% masking probability maintain a steeper slope, meaning that the bigger the unmasked context, the better-generated variants, as expected.

### 2.2.3 Knowledge-Guided Optimization in Action

To fulfill the demonstrated utmost computational efficiency coupled with the top downstream performance, our model design was preceded with knowledge-guided experimentation. We define knowledge-guided experimentation as protein-specific experimentation retaining a single independent variable traversing masking (strategy and probability), architecture, and pre-training dataset. This pre-design experimentation retained training each variation for 2 epochs while also abiding by approximately the same total number of parameters per experiment to avoid complexity biases. The definitions of the experimental

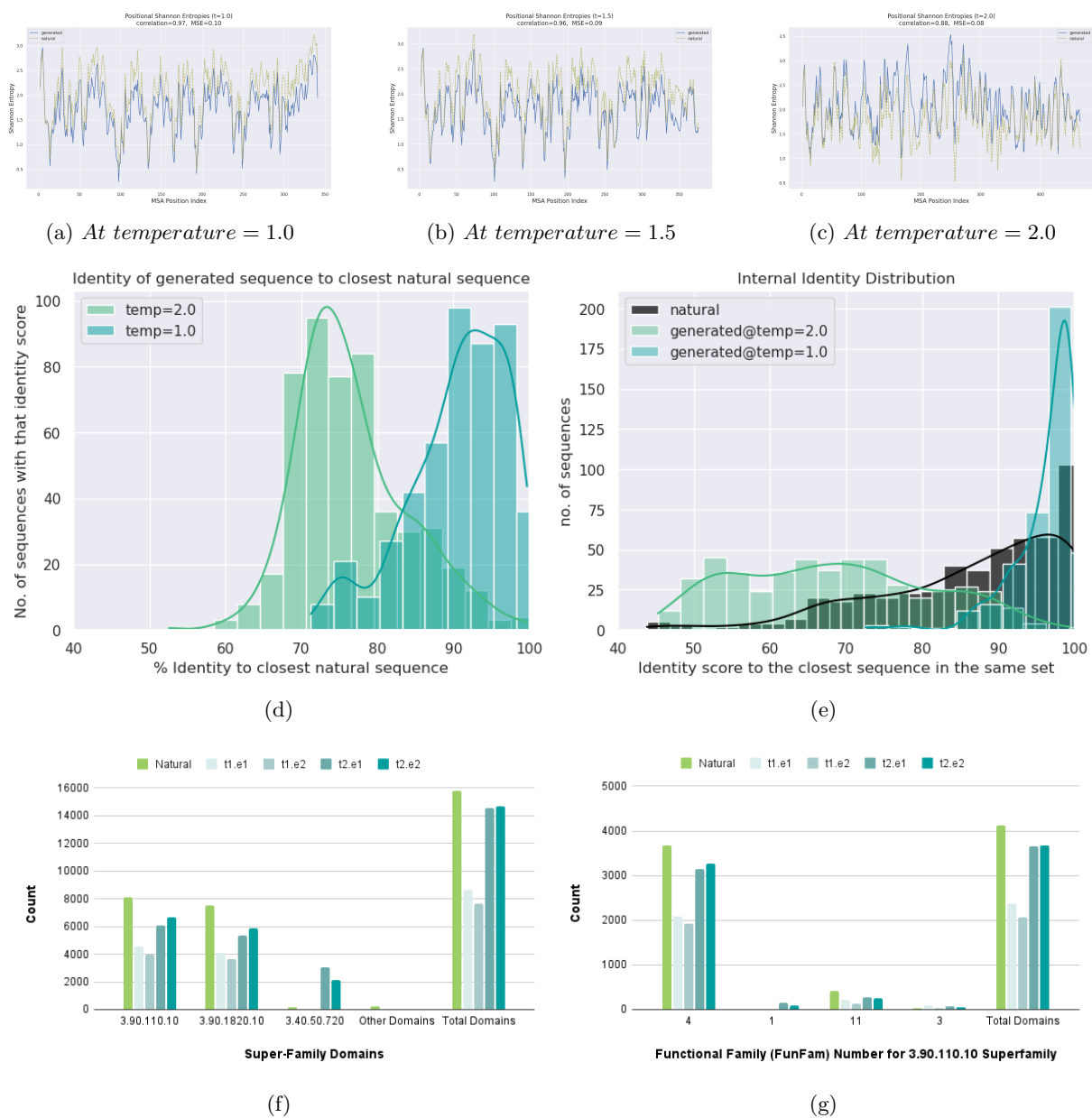

(a) *At temperature = 1.0*  (b) *At temperature = 1.5*  (c) *At temperature = 2.0*

(d)

(e)

(f)

(g)

**Fig. 2**: **Auto-Regressive Fine-Tuning Generation of MDH variants:** (a), (b), and (c) Shannon entropy curves between the MSA of natural and generated sets at logit warping temperatures of 1.0, 1.5, and 2.0, respectively: Manipulating the temperature during the generation affects the variability of the generated sequences. (d) Distribution of the Sequence Identity of the generated variants of temperatures 1.0 and 2.0 vs. natural variants: The generated sequences span a wide range of sequence identity scores going as low as 60%. (e) Internal identities of the natural/fine-tuning set vs. the generated sets of temperatures of 1.0 and 2.0: The generated variants span wide ranges of variability, with the set of temperature 2.0 showing the widest range. (f)CATH domain distribution of the three dominating homologous super-families in the natural set: All the generated variants retained CATH domains from the three dominating super-families. (g) The distribution of CATH domains with known functional annotations in the 3.90.110.10 super-family (responsible of the dehydrogenase function) in the natural and generated sets: The generated sets managed to maintain domains from the three functional families of the natural set while expanding to domains from functional family number 1, maltase dehydrogenase (MDH), with only 6 examples in the natural set. Abbr., Nat: Natural sequences; t1.e1: generated sequences at temperature 1 by the first epoch; t1.e2: Generated sequences at temperature 1 by the second epoch; t2.e1: Generated sequences at temperature 2 by the first epoch; t2.e2: Generated sequences at temperature 2 by the second epoch

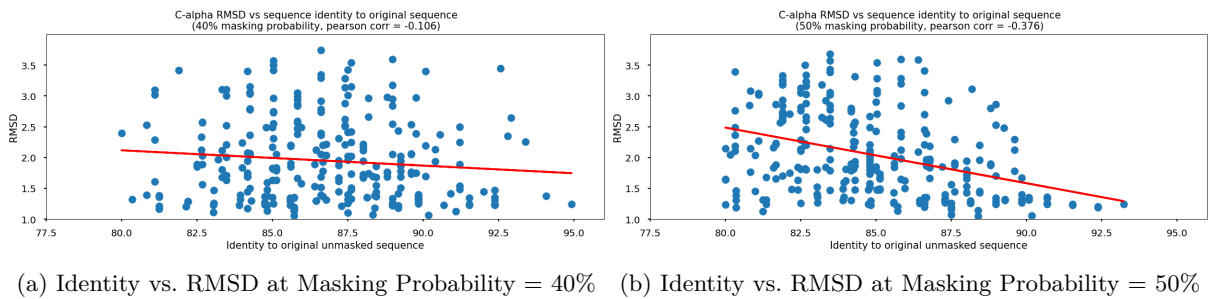

(a) Identity vs. RMSD at Masking Probability = 40%    (b) Identity vs. RMSD at Masking Probability = 50%

**Fig. 3**: **Sequence Identities vs Structural Similarities of the Generated Sequences at Masking Probability of 40% (Fig 3.a.) and 50% (Fig 3.b.):** The figures show the model was able to generate sequences with RMSD lower than 1.5 A withe sequence identity as low as 80% in both cases. The generated sequences at masking probability= 50% show more negative Pearson correlation, suggesting more sequence context facilitates the generation of similar structures with low sequence identities.

independent variables, their values, evaluation metrics, and the experimentation baseline can be found in the Methods section. The design of the final models integrates the top-performing experimental versions of each independent variable's set of experiments and sub-experiments. We fully pre-trained two models, *Ankh* and *Ankh_base*, that can be visualized in Figure *4*. The results of the masking strategy experiments promoted utilizing a 1-gram span de-masking strategy (elaborated on in *Exp. 4*). Regarding the masking probability, we used 20% as promoted by *Exp. 8*. For the number of layers, we chose 48 layers for the encoder and 24 layers for the decoder as per the results of *Exp. 11*. For the activation function, we proceeded with Gated-GELU as per the findings of *Exp. 14* and *Exp. 15*. We adopted relative positional embeddings with an embedding offset of 128 and an embedding dimension of 64 as per the results of *Exp. 20*. However, the only dimensions where the two models differ are the embedding dimension, number of attention heads, and number of feed-forward layers. *Ankh_base* has an embedding dimension of 768 as recommended by *Exp. 13* while *Ankh* has an embedding of 1536 as we found that double dimensions are often better performing as verified by *Exp. 20*. Furthermore, the number of attention heads for *Ankh_base* and *Ankh* is 12 and 16, respectively. Finally, the number of feed-forward layers for *Ankh_base* is 3072 and for *Ankh* is 3840. The full configurations for the two models can be found in Table *11*.

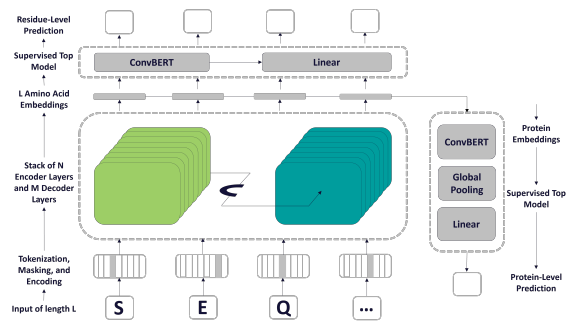

**Fig. 4**: **Architecture of Ankh Models**: Arrows show the information flow in the network starting from the input sequences, transformer input pre-processing, transformer, and then either a residue-level prediction network or a protein-level prediction network that only differs in being preceded by a global max pooling layer. Both *Ankh_base* and *Ankh_large* agree on the demonstrated architecture. However, they differ in the length of the context vector.

## 3 Discussion

**Results Summary - More Efficient Models Can Also Generalize Better:** To fulfill a holistic analysis of our model, we ensured an evaluation of Ankh that spans the most principal and valuable categories of protein modeling via deep learning. From a deep learning perspective, we have tested our model on regression, classification, and generation. From a biology perspective, we have tested our model on residue-level and protein-level where the measured attributes spanned structure and function. From a scale and sparsity perspective, we have tested our model on both High-N and One-N scales. In all of the aforementioned benchmarks, Ankh surpassed the performance of state-of-the-art with no exceptions while our base model Ankh base either reached the same or achieved very comparable with significantly less computational power, offering two modes of computation depending on the application demands. In our generation analysis, Ankh was able to learn the evolutionary conservation-mutation trends and introduce diversity while retaining key structural and functional characteristics in both High-N and One-N scales.

**Results Interpretation:** Our results reinforce the compatibility of PLMs with the language of life but signify protein knowledge-guided model design. Moreover, our results fortify the use of sequence contextualized embeddings as a mere input to downstream models but shed light on task-specific architectures/layers.

**Results Implications:** Our results suggest that state-of-the-art performance can be reached and surpassed by significantly less computational power. This suggestion implies the necessity of criticizing the highlighted correlation between model performance and needed computational power, embodied in either model or data sizes. Instead, our work suggests visualizing this correlation as a trade-off, highlighting the immense cost of directly scaling model/data size to improve model performance. On the other side of the trade-off, we propose knowledge-guided means of optimization whose prerequisites revolve around needed protein knowledge as well as the extra mile in optimizing both the software and hardware components of the model life cycle.

**Results Limitation:** We report several limitations of our work. Firstly, changes of the activation function impacted the optima in the number of layers for encoder and decoder, as well as, the embedding dimension. However, this setting is traced to the nature of the Gated-GELU activation function denoting a significantly larger number of trainable parameters that, in turn, forced us to compensate increases in width by decreases in depth to retain the same total number of trainable parameters and avoid any possible complexity-bias [15]. Consequently, testing different combinations of dimensionality required utilizing an activation function needing fewer parameters. Secondly, our optimization propagated the top-performing model version for each prediction task to the next one. *Top performance* was not achieved through the version numerically performing top in all tasks, because none such version existed. Instead, we selected the version that outperformed others for most of the most standard data sets, and performed top for diverse objectives. The downside of this seemingly simple algorithm is that we failed to define any single formula to compute "best" that stands out from amongst the population of such formulas. Specifically, we documented the reasoning behind the choice of the top performer for each prediction task. Generally, we justified our choices by opting for holistic predictions via a general-purpose language model. This entailed promoting generalization across different tasks, especially when the difference amongst results remained numerically negligible, when acknowledging that task-specific customization is of impact, and when all the experimental versions are only trained for two epochs.

**Recommendations:** Our results implicitly suggested that the choice of the data used to pre-train the pLMs might have to be coordinated with that of data sets used for testing subsequent protein prediction tasks. Although such an endeavor is beyond this work's scope, we encourage efforts toward this end. For instance, we have reported the superiority of pre-training with UniRef50 over UniRef90, UniRef100, and BFD due to lower redundancy. Although the details of what exactly constitutes redundancy relate to the application (e.g. using all available human proteins constitutes less redundancy when wondering about the length distribution of human proteins than when trying to predict binding residues in these), too much redundancy is often easy to spot [16]. Furthermore, we shed light on the value of incorporating synthetically-generated-experimentally-characterized sequences.

**Future Work:** We present Ankh as an initial version of our optimized general-purpose protein language model. This version is meant to serve as a pre-training base that shall then be specialized into high-impact and high-value protein modeling tasks in our future work (e.g., full atomic resolution 3D-structure prediction, protein generation, etc.) with task-specific optimization and detailed analysis.

# 4 Methods

## 4.1 Self-Supervised Learning

### 4.1.1 Pre-training Datasets

Deep learning-based protein modeling, likewise NLP, is data-driven. However, this hunger for data is proving to be both constrained and constraining. We draw upon previous experimentation done in ProtTrans, in general, and Prot-T5, in particular, where the performance and associated computational power of three datasets ranging in size, identity-based clustering, and origin were analyzed. The analysis promoted utilizing UniRef50 [17] over UniRef100 [17] and BFD [3, 18]. We build upon the same results by pre-training our baseline on UniRef50.

UniRef (UniProt Reference Clusters) databases offer variable clustering of UniProtKB sequences (including isoforms) and selected UniParc records [17]. This variable clustering denotes different sequence similarity thresholds that can fulfill non-redundancy and intra-cluster homogeneity. In UniRef100, a single cluster/entry denotes identical sequences and sub-fragments originating from any organism. To build UniRef90 and UniRef50, UniRef100 sequences are clustered at 90% and 50% sequence identity thresholds, respectively [17]. Hence, the results in ProtTrans can be traced to having more variability and representation in UniRef50 [3]. Nevertheless, we wanted to test an intermediate value between the 50% and 100% thresholds that was not tested in ProtTrans, hence the justification for having the pre-training dataset amongst our experimentation independent variables later on with a value that is, UniRef90. We can refer to the pre-training data statistics in Table 2 Sequences in both UniRef50 and UniRef90 are tokenized with a single space between each token in analogy to word boundaries and each sequence is stored on a separate line in analogy to sentences.

### 4.1.2 Pre-trained Model: Encoder-Decoder Transformer

For our baseline model and throughout our experimentation, we utilize the encoder-decoder transformer originally proposed for machine translation and then for mapping an arbitrary input domain to a target domain [19]. The encoder learns to project the input domain sequences into a latent/embedding space, representing the "context". The decoder, however, learns to generate the target domain sequences given this context. Although later transformer releases abandoned the encoder-decoder combination and only utilized either of them, we draw upon Prot-Trans's experimentation promoting T5 [20] (the only encoder-decoder transformer analyzed) over encoder-only (e.g., BERT [21], ALBERT [22], and Electra [23]) and decoder-only transformers (e.g., TransformerXL [24] and XLNet [25]). However, the choice of encoder-decoder transformers in this work is not only motivated by the aforementioned top performance but also their compatibility with our experimentation independent variables such as, but not limited to, masking and architecture. Due to retaining both the encoder and decoder, T5-like transformers offer more compatibility with different masking strategies that are dependent on both masking and de-masking techniques. Furthermore and due to learning relative positional embeddings that are shared across layers for each attention head, T5-like transformers offer robustness against predictions surpassing the maximum length of the relative positional embedding dimension as it learns to combine the relative offset between lower layers' amino acid subsets [20]. For the masking strategy adopted by the baseline, we performed a 1-gram random token masking according to the default probability of 15% (i.e., 15% of the sequence tokens are masked at random) and performed a full de-masking/reconstruction of the sequence (i.e., all tokens are reconstructed as individual tokens). For the number of encoder-decoder layers, we used 36 layers for each. For the activation function, we used Gated Gaussian Error Linear Unit (Gated-GELU) [15]. Regarding relative positional embedding, we have been using an embedding dimension of 32 and an embedding offset of 128. Finally, we pre-trained the baseline on UniRef50. The remainder of the baseline configurations can be found in Table 11. Our baseline model represents *Exp. 0* that is later subjected to a single variable change at every class of experimentation.

## 4.2 Downstream Tasks

To provide a holistic analysis of Ankh's performance -relatively and ultimately-, we conduct a protein downstream benchmark as well as a generation analysis for protein engineering applications. For the protein downstream benchmark, we measure the performance of Ankh in comparison to the top reported protein language models via selected downstream tasks covering various aspects of protein understanding and involving protein structure and function. For the generation analysis, we analyze the use of Ankh via two generation frameworks on two scales of data: High-N (Family-based) and One-Shot (single sequence-based) generation in terms of conservation-mutation trends, introducing diversity while retaining structural -and accordingly functional- identity, and sparsity-robustness. We unified the experimental training and testing settings and procedures of all the downstream tasks investigated in this study. Although we acknowledge task-specific optimization, this unification aims to specifically compare the level of protein understanding embodied in the protein representations generated by the studied models while avoiding any bias that can result from task-specific means of optimization. We can observe a demonstration of the adopted downstream tasks in Figure 5. The following subsections are dedicated to the description of the tasks, task databases, and the predictive models/settings that utilized them.

### 4.2.1 Tasks and Datasets

The emergence of transformers as powerful self-surprised models for proteins has motivated significant efforts in designing comprehensive benchmarking databases for protein sequence understanding, like TAPE [26], and PEER [27]. These efforts are necessary to drive the progress of transformers toward protein understanding, parallel to how comprehensive benchmarks have driven the progress of transformers in natural language understanding. We provide a summary of the downstream dataset statistics in Table 12. We selected a set of commonly-utilized downstream tasks that fall into three groups: Protein Function Prediction, Protein Structure Prediction, and Protein Localization Prediction. We further extended the validity of the comparison by adding independent testing databases to some of the downstream tasks. Besides the benchmarking tasks, we also investigated the protein variant generation capabilities of Ankh in two settings: protein family-based generation and single protein-based generation.

### 4.2.2 Protein Function Prediction

This group of tasks aims to evaluate the ability of protein embeddings in capturing the functional scores of two critical design parameters of protein engineering, fluorescence, and solubility.

**Fluorescence Prediction (FluP):** This regression task evaluates the fluorescence intensity of green fluorescent protein mutants annotated by Sarkisyan et al [28]. Fluorescence is an important biological function as it allows researchers to infer the existence of proteins in cell lines and living organisms [27]. Prediction of the effect of mutations on the green fluorescent protein is a common example to investigate, representing the protein genotype/phenotype fitness landscape prediction problem in protein engineering research. We follow the same splits of TAPE [26] as they are designed to test the model generalization ability from lower-order mutations to higher-order mutations. The training and evaluation datasets contain only mutants with three or fewer mutations while the testing contains mutants with four mutations or more.

**Solubility Prediction (SolP):** This classification task evaluates the binary label of a set of dissimilar proteins as soluble or not. Solubility is an indispensable design parameter for effective proteins, especially for pharmaceuticals [29]. We adopted the solubility database utilized in the development of DeepSol [30] with the same dataset splits, where any protein with $\geq 30\%$ sequences identity to any protein in the testing subset is removed from the training and evaluation subsets.

**GB1 Fitness Prediction (GB1):** This regression task evaluates the fitness of GB1 binding following mutations in 4 specific positions curated in the FLIP benchmark [31]. GB1 is the binding domain of the immunoglobulin-binding Protein G, used in antibody purification. GB1 landscape is considered the gold standard in studying the non-additive interactions of mutations, termed epistasis [32]. Unlike the fluorescence mutants, the GB1 mutants are confined only to four positions, which is why it is selected to represent another common case in protein engineering research. In FLIP, 149,361 GB1 mutants with measured binding scores were down-sampled to 8,733 as 96% of the original data

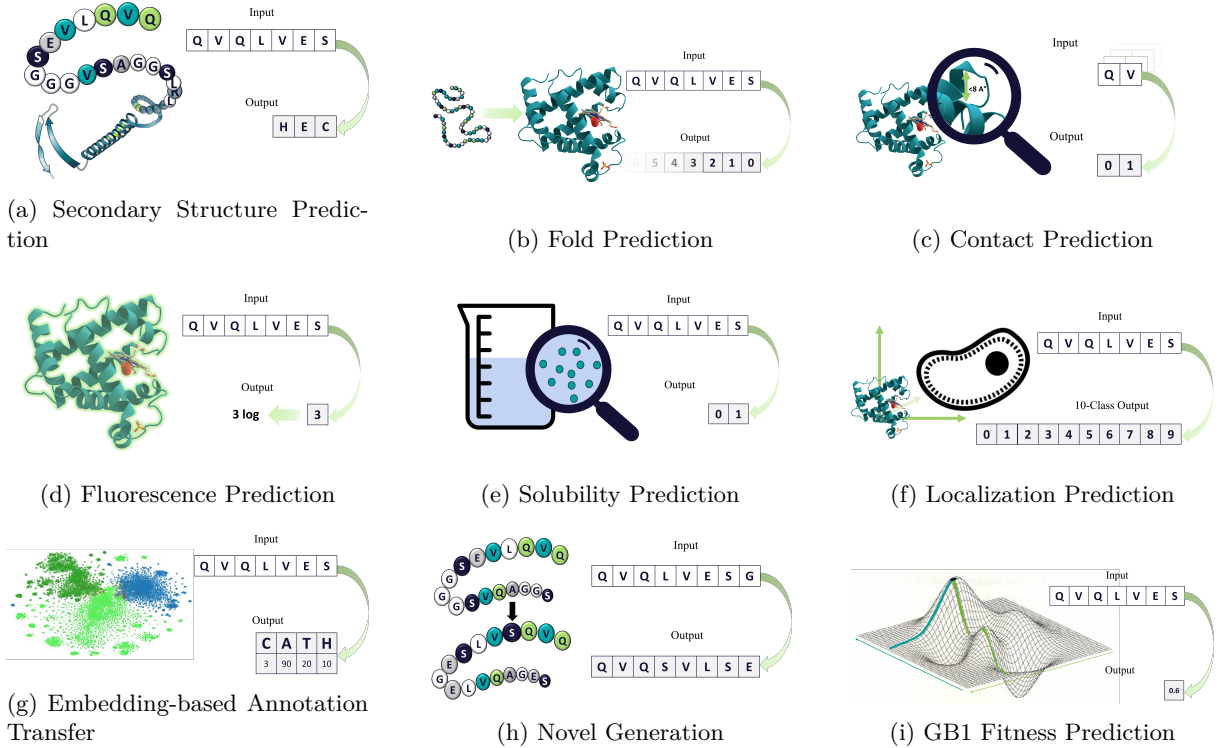

(a) Secondary Structure Prediction

(b) Fold Prediction

(c) Contact Prediction

(d) Fluorescence Prediction

(e) Solubility Prediction

(f) Localization Prediction

(g) Embedding-based Annotation Transfer

(h) Novel Generation

(i) GB1 Fitness Prediction

**Fig. 5**: A Demonstration of Structure & Function Benchmarks: We showcase all the adopted downstream tasks (a) In SSP, the input is the protein sequence and the output is a per-residue classification that spans either 3 or 8 states. (b) In FolP, the input is the protein sequence and the output is a per-protein classification that spans 1194 possible folds. In CP, the input is the protein sequence that is processed as residue pairs and the output is a binary classification indicating whether or not the designated residues contact. (d) In FluP, the input is the protein sequence and the output is a regression score indicating the fluorescence intensity. (e) In SolP, the input is the protein sequence and the output is a binary classification indicating whether or not the protein is soluble. (f) In LocP, the input is the protein sequence and the output is a per-protein classification that spans 10 classes. (g) In EAT, the input is annotated query protein and the output is CATH annotation transferred from the best match in annotated lookup dataset (h) In Novel Generation, the input is a protein sequence and the output is a variant of the input protein that maintains the same desired function. (i) In GB1 Fitness Prediction, the input is a protein sequence corresponding to mutations at four possible positions and the output is a regression score indicating the fitness prediction

points were non-binder or poor-binder. The sampled split was utilized to evaluate the investigated PLMs as it provided the most stable results.

### 4.2.3 Protein Structure Prediction

This group of tasks aims to evaluate the ability of the sequence-based embeddings of a protein to encompass accurate information about its structure. A sequential chain of amino acids folds into a set of predetermined stable three-dimensional structures. The large majority of biological parameters of a protein can be inferred from its structure [33]. Consequently, this group of tasks is esteemed due to its correlation to protein understanding, and the large set of applications it enables.

**Contact Prediction (CP):** This is a binary classification task, where pairs of residues are predicted to be in contact in their 3D structure (commonly defined with an 8Å distance threshold) or not. Contact prediction, as a residue-level 3D structure prediction, provides significant global information about the protein structure. In literature, contact prediction is utilized as an intermediate prediction step toward atom-level 3D structure prediction. We utilized ProteinNet [34], a standardized dataset for structure learning, whose approach is to piggyback on the CASP competitions [35]. ProteinNet uses the CASP structures as a testing set and augments all the historical records of structures released before the CASP dates as training and evaluation sets. We utilized the latest version of ProteinNet that uses CASP12 as its test set with the same dataset splits as TAPE [26]. To further asset the robustness of the model training, we add independent testing based on the free modeling (FM) structures of the latest CASP competition, CASP14.

**Fold Prediction (FolP):** This is a classification task, where a full protein sequence is classified into 1194 possible folds. This task is utilized in the detection of emergent remote homologs of proteins of interest like new antibiotic-resistant genes, and industrial enzymes [36]. We adopted from Hou's dataset [37], with its original splits. Entire clustered superfamilies are held for the testing dataset to affirm the models' generalization ability to detect the structural similarity of drastically different sequences.

**Secondary Structure Prediction (SSP):** This is a classification task, where each residue in a protein is classified into its secondary structure fold with two levels of difficulty: 3-classes, and 8-classes. Secondary structures hold significant information about functional domains and are commonly utilized to capture evolutionary information through multiple sequence alignment. We utilized the training and evaluation set from NetSurfP-2.0 [38] and used a variety of testing sets to affirm the robustness of the model including CB513 [39], TS115 [40], and CASP12 [41], and CASP14 [42].

**Embedding-based Annotation Transfer (EAT):** Protein annotation transfer from labeled (experimentally-annotated) proteins to unlabeled proteins traditionally employed Homology-based inference (HBI) in sequence space. Recently, embedding-based annotation transfer has emerged as an alternative faster approach, as it does not require multiple sequence alignment (MSA) calculations. In the new framework, the distances between query proteins and a lookup set of annotated proteins are calculated to transfer annotation from the most similar known match through k-Nearest Neighbors (k-NN) in embedding space [43, 44]. This task evaluates the ability of raw embeddings to capture meaningful information about the proteins without the need for supervised training. We utilized a bench-marking set of 69k look-up sequences and 219 test sequences that were developed for ProTucke evaluation [44]. This evaluation dataset is curated from CATH v4.3 dataset, where proteins are classified into four levels of structural annotations: Class, Architecture, Topology, and Homologous super-family [45]. For simplicity, we report the mean of the four accuracy scores as a performance measure for this task in Table. 1. The performance over the four classes was consistent with the mean as seen in Table 16.

### 4.2.4 Protein Localization Prediction

This task aims to evaluate the ability of protein embeddings to capture where a protein is expected to accumulate in the cell, known as protein localization prediction [46]. This attribute is significant for understanding protein functions, especially in disease target identification studies.

**Sub-cellular localization prediction (LocP):** This classification task evaluates the localization of a protein into 10 sub-cellular classes. We utilize the DeepLoc [47] dataset with the same dataset splits described in their paper.

### 4.2.5 Generation of Novel Protein Sequences

Following the comparative study, we utilized the *Ankh* model to generate synthetic variants of natural proteins to affirm its applicability in a crucial protein engineering task, computational variant generation. We evaluate Ankh with two input datasets representing two different settings and scales in protein engineering: family-based and single sequence-based variant generation.

**High-N (Family-Based Variant Generation):** For the family-based use case, we utilized a curated dataset of malate dehydrogenase (MDH), which was utilized as the training dataset of ProteinGAN [48]. ProteinGAN is a recent deep learning-based generative model that showed superior performance compared with experimental variant generation in the case of MDH. This choice of this protein family is convenient given its diverse 16,706 unique training sequences as well as the complexity of enzyme catalysis as it requires binding to both its substrate and the NAD+ cofactor, which add complexity to the generation process

**One-Shot (Single Sequence Variant Generation):** For the single sequence use case, we used single chain SARS-Cov-2 nanobody that was added to the CoV-AbDab dataset after June 2022 [49]. This is to ensure that they are new sequences the model did not see in its unsupervised training. This one-shot generation use case specifically challenges the model's generalization capability by demanding it to generate variants of a small-scale dataset without over-fitting on the dataset in question. We conducted seven independent virtual generation experiments, utilizing seven different nanobodies Cov-AbDab identified by the following names: Nb-007, F6, Nb_1-23, Nb_1-25, Nb_2-62, Nb_2-65, and Nb_2-67. All of the selected nanobodies have experimentally validated structures to facilitate their comparison with the predicted structures of the generated candidate nanobodies.

## 4.3 Downstream Model: ConvBERT

For our implementation of the top/supervised models mapping the pre-trained embeddings to the designated supervised targets, we utilized the same supervised network with very few modifications to account for the differences in protein processing levels (e.g., residue-level and entire protein-level) and output distributions (e.g., binary classification, multi-class classification, and regression). In all cases, our top/supervised models consist of two types of layers. The choice of the first type draws-upon previous experimentation done in ProtTrans promoting CNNs as top/downstream models/layers that are proven to perform better when coupled with self-attention [3]. We utilize a ConvBERT layer with the same embedding dimension as the pre-trained model, a feed-forward network dimension of pre-trained embeddings divided by 2 (i.e., if we're benchmarking with ESM-1b whose embedding dimension is 1280 then the feed-forward network dimension will be 640), 4 attention heads, a dropout rate of 0.2, convolutional kernel size of 7, and a Gated-GELU activation [50]. The second type is linear layers whose activation varies between None, Sigmoid, and Softmax for regression, binary classification, and multi-class classification, respectively.

The third type of layers, however, is not shared across all cases. In fact, we only used a global max pooling layer only in regression and binary classification tasks prior to/at the beginning of the aforementioned supervised network. Generally, we acknowledge -and promote- task-specific optimization. Accordingly, we acknowledge that different top models with different set of hyperparameters and configurations can result in a better downstream performance. Nevertheless, we unify the setting of this top model believed to achieve the better generalized performance as the core of the downstream bench-marking is to evaluate and compare the level of protein understanding embodied each model's learned protein representations.

## 4.4 Variant Generation Model

The proposed auto-regressive fine-tuning generation framework for the High-N (protein family-based generation) scale offers an accessible approach for variant generation, that can be easily scaled across different protein families. Moreover, the framework offers easy control over the exploration-exploitation trade-off of the generation, by manipulating the logit warping temperature parameter. To validate the framework, the same fine-tuned model was first utilized to generate three datasets of 500 sequences, utilizing three different temperatures (1.0, 1.5, and 2.0). For the three datasets, the Shannon entropy curves between the generated set and a multi-sequence alignment (MSA)of the natural set are reported. Shannon Entropy aims to characterize if the generated sets preserve the evolutionary sequence properties of the natural MDH set by comparing their representative statistics of amino acid variation. Low entropy values reflect conserved residues governing retained functionality whereas high entropy values reflect less conserved regions with higher mutation rates. We can observe in Figure 2, the three generated sets show high similarity to the distribution of the natural sequences with almost identical positions for both peaks and valleys. MSE between entropy values of natural and generated sets are quantified as 0.1, 0.09, and 0.08 for a generation with temperatures 1.0, 1,5, and 2.0 respectively. We emphasize that the reported similarity is calculated based on generated set of 500, which is less than 3% of the total sequences of the natural set (16,700). In other words, the model is able to mimic the distribution of the natural set with a small portion of its original sequences.

Since variant generation tasks adopt a framework that is different from the aforementioned pretrained-top model framework and evaluation settings, we present it solely. We adopt two frameworks for variant generation, Auto-Regressive Fine-Tuning for High-N (family-based generation) and Masked Language Modeling (MLM) for One-Shot (single protein-based generation).

Fine-tuning denotes specializing the pre-trained model on a specific dataset/task(s) [51]. We define auto-regressive fine-tuning, however, by adding a constraint to the classic fine-tuning setting that denotes completely freezing the encoder and only allowing for the decoder's parameter to change. We allow this for the entirety of the decoder's layers and initialize the fine-tuning by the original decoder parameters. We train each experiment for 2 epochs, shifting it from an masked language modeling prediction into an auto-regressive prediction. We set a maximum sequence length of 256 tokens and a maximum prompt length of 20 tokens. We utilize a learning rate of $3e-4$ and an epsilon value of $e-8$ for the Adam optimizer. Moreover, we use a train batch size of 4 and an evaluation batch size of 8. For the auto-regressive sampling, we use beam search with a number of beams of 10. For the auto-regressive logit warping, we use temperature with three different temperature values of 1, 1.5, and 2 to observe the behavior of the model under different temperatures. The mixture of beam search sampling and temperature logit warping works as follows: Firstly, temperature changes the logit distribution, preserving the order of the tokens but smoothing/sharpening the distribution. Secondly, sampling occurs where the beams are scored (greedily, in our case).

MLM denotes a pre-training objective that guides the model's learning -and accordingly its inference- of token representations by requiring it to predict a random sample of input tokens that is usually replaced by a $[MASK]$ placeholder(s) in a multi-class setting over the entire vocabulary [51]. However and for fine-tuning purposes, we use MLM in the context of inference only (i.e., no training or change of parameters of any kind has been done). Furthermore and unlike the commonly-used small values of masking probabilities for pre-training purposes, we perform two experiments where we use larger masking probabilities corresponding to a bigger range of variations in the original sequences. The two experiments try different variations of the exploration-exploitation trade-offs. The first experiment attempted a higher masking probability of 50% and a lower temperature logit warping of 1.0. The second experiment utilized a lower masking probability of 40% and a higher temperature logit warping of 1.5. Both experiments utilized a beam search sampling with 30 beams.

## 4.5 Computational Power (Software & Hardware)

Processing a limited vocabulary that incorporates unlimited life within its tokens, we had to also reach out for the limits of up-scaling the efficiency of our computational power in terms of both software and hardware.

### 4.5.1 Flax & JAX

Flax is an end-to-end high-performance library and an ecosystem for JAX that is designed for flexibility and tailored for neural networks. JAX opts for a plausible range of composable function transformations, allowing just-in-time (JIT) compilation, automatic differentiation, CPU/GPU/TPU compatibility, and automatic vectorization and parallelization [52]. JAX, for example, can deliver approximately 1.4x speed-up for language model training on TPU Pods compared to PyTorch. We used Flax to build our source codes to leverage the aforementioned advances during training the PLMs on Google TPU Pods but made our models available on HuggingFace to support a wider range of researchers, which use any of the three top deep learning libraries: JAX, TensorFlow, and PyTorch.

### 4.5.2 TPUs

We were fortunate enough to be amongst the first manuscripts, in general, and to be the first protein modeling work, in particular, utilizing Google's latest TPU v4 and unleashing unseen capabilities of supercomputers. This means all the pre-trained models on this work were trained using Google

TPU v4 Pods with either 64 or 128 cores. The single TPU v4 VM host has 8 TPU cores, and each core has 16 GiB of high-bandwidth memory, 120 CPU cores, and 400 GB of main memory. At first glance, it seems TPU V4 is similar to TPU V3; however, it has two main advantages over TPU V3. First, the new mega core features allow the virtual merge of 2 cores, making deep learning libraries like JAX sees every two cores as one core. This gives each of the four merged cores per host access to 32 GiB of memory, which leads to fitting bigger models up to 3 billion on a single host without the need to use model parallelism. Second, TPU V4 can deliver approximately 2.2x speedup compared to TPU V3.

## 4.6 Data & Model Experimentation

### 4.6.1 Masking

Masking is a pre-training objective that guides the model's learning of token representations by demanding it to predict a random sample of input tokens that is usually replaced by a $[MASK]$ placeholder(s) in a multi-class setting over the entire vocabulary [21]. This class of model experimentation aimed to investigate the impact of two masking-related parameters, masking strategy and masking probability.

**a. Masking Strategy:** Masking strategy indicates the means by which we decide which tokens to mask and which to keep unmasked [53–55]. Motivated by the skewed distribution of amino acid tokens in protein sequences in addition to the redundancy in the database, we have tested different means of masking strategies to ensure protein-specific adoption. For that purpose, we tested out five variations/experiments.

**Exp. 0:** This experiment represents our baseline model whose details can be found in Sub-Section 3.1.2.

**Exp. 1:** We masked every unique 1-gram token (i.e., every unique amino acid) in the sequence at least once. We did so by repeatedly iterating the sequence and randomly masking an amino acid at a time providing that neither the desired masking probability has yet been reached nor that the amino acid in question is the one with the highest count (i.e., if we are using a 15% masking probability for the sequence "$ABCAAAAAAA....A$" whose length is 20 amino acids: token "$A$" will always be masked one time and the remaining two masks will always be for tokens "$B$" and "$C$"). Finally, we performed a full de-masking/reconstruction of the whole sequence tokens. This has been found to achieve higher performance w.r.t. the baseline and accordingly was proceeded with.

**Exp. 2:** We have also masked every unique token in the sequence at least once but instead of masking one token at a time, we masked its precedent and subsequent tokens (i.e., in the sequence "$ABCDEFG$": if token "$D$" is masked, then tokens "$C$" and "$E$" are also masked), turning it from a 1-gram token masking to 3-gram token masking. We also retained the full de-masking/reconstruction of the whole sequence tokens. This, however, reduced the performance in all the downstream tasks and accordingly was not proceeded with.

**Exp. 3:** We replicated *Exp. 1* indicating the masking of every unique 1-gram token in the sequence at least once. However, as opposed to the default scenario where all tokens' de-masking/reconstruction is incorporated in the calculation of the loss function, we have only incorporated the reconstruction of the masked tokens even though the output will contain the entirety of the sequence tokens (i.e., in the sequence "$ABCDEFG$": if tokens "$D$" and "$F$" are masked, then only the reconstruction of tokens "$D$" and "$F$" is accounted for the calculation of the loss function). However, this reduced the performance in all the downstream tasks and deducing that we should reconstruct the entire input, even if it is already known, and accordingly was not proceeded with.

**Exp. 4:** This experiment reflected a change in the input-target mapping of masking and de-masking. To elaborate, every input token is reconstructed as a single target token in the default 1-gram token masking case. The change we performed was to reconstruct all the consecutive unmasked tokens as a single merged token (i.e., if the input sequence was "$ABCDEFG$" and tokens "$C$" and "$G$" were masked then the sequence was inputted as "$A, B, [MASK_0], D, E, F, [MASK_1]$" and reconstructed as "$[MASK_0], C, [MASK_1], G$" where $[MASK_0]$ is a single target token mapping two input tokens (AB) and $[MASK_1]$ is a single target token mapping three input tokens (DEF), turning it from a 1-gram token masking into a 1-gram span masking. The change was motivated by reducing the computational cost of the unneeded computations associated with the reconstruction of unmasked tokens of the output. We refer to this masking strategy as "1-Gram Span Partial Demasking/Reconstruction". It is important to note that this partial reconstruction is done only on the output as the input tokens are left as is. This direction has proven to be a valid one, corresponding to higher performance w.r.t. the first experiment and accordingly was proceeded with for the upcoming experiments.

**Exp. 5:** We applied the 1-gram span partial reconstruction -introduced in *Exp. 4*- on the approach of *Exp. 1*, indicating the masking of every unique 1-gram token in the sequence at least once and the reconstruction of all the unmasked tokens as a single merged token (i.e., if we are using a 15% masking probability for the sequence "$ABCAAAAAAA....A$" whose length is 20 amino acids and if the first random index was the zeroth index then the sequence will be inputted as "$[MASK_0], [MASK_1], [MASK_2], A, A, A, A, ...., A$" and reconstructed as "$A, B, C, [MASK_0]$" where "$[MASK_0]$" is a single token mapping seventeen input tokens). This, however, was shown to reduce the performance, consistently, and was therefore discarded.

**Exp. 6:** We have tried a variant of *Exp. 4* in terms of the partial reconstruction where we mapped all the consecutively-masked tokens into a single token upon reconstruction (i.e., if the input sequence was "$ABCDEFG$" and the tokens "$C$", "$D$", and "$E$" were masked then the sequence was inputted as "$A, B, [MASK_0], F, G$" and reconstructed as "$[MASK_0], [MASK_1], [MASK_2]$", where "$[MASK_1]$" is a single token mapping three masked tokens, $[MASK_0]$ and $[MASK_2]$ are single tokens mapping two unmasked tokens) turning it into a 3-gram span masking. This change, too, was motivated by reducing the computational cost but has shown to be an invalid direction and accordingly was not proceeded with.

Hence, it can be deduced that the top performing version of the six tested versions was the version of *Exp. 4*, where we reconstruct all the consecutive unmasked tokens as a single merged token. Therefore, this was the version that carried on to the following sub-set of experiments, masking probability. The results for this set of experiments can be found in Table 3.

**b. Masking Probability** Masking probability indicates the ratio of tokens to be masked out of the entire sequence length. As indicated in the baseline model's configurations, the default masking probability is 15% [21]. Here, we have experimented with three additional values on the top-performing version of *Exp. 4*.

**Exp. 7:** The first tested masking probability was 10%.

**Exp. 8:** The second tested masking probability was 20%.

**Exp. 9:** The third tested masking probability was 30%.

Out of the four values, 10% was the worst masking probability and was accordingly disregarded. Interestingly, it was found that the default value of *Exp. 4* (15% masking probability) was outperforming for the sub-cellular localization prediction, fold prediction, as well as some of the secondary structure prediction tasks for datasets such as *CB513* and TS115. Nevertheless, it was found that the value of *Exp. 9* (30% masking probability) was outperforming for the entirety of the contact prediction tasks in addition to the secondary structure prediction tasks for CASP12 dataset. Given the inconsistency amongst the secondary structure prediction dataset results, we referred to the results on CASP12 -being a domain standard- that promotes the higher masking probability. Furthermore and as we are opting for holistic predictions via a general-purpose language model, we promoted generalization across different types of tasks, especially when the difference amongst results is of such a small magnitude and when acknowledging that task-specific customization is of impact. Finally, given that all the experimental variations were trained for only two epochs, we proceeded with the intermediate value of *Exp. 8* (20% masking probability) for the post-experimentation long-term training to fulfill the semi-comprehensive inclusion of different polls of tasks anticipated from a general-purpose protein language model that can then be customized per downstream task. The results for this set of experiments can be found in Table 4.

### 4.6.2 Architecture

Since we are utilizing an encoder-decoder transformer architecture, the architecture variations we target correspond to the number of encoder and decoder layers, different combinations of depth and width variations, and the means by which the model learns the order of tokens (i.e., positional embeddings).

**a. Number of Encoder-Decoder layers**
Although the presence of a decoder is essential in improving the representations produced by the encoder, previous runs of Prot-T5, deduced that the decoder did not provide any notable difference in most of the downstream tasks and was accordingly eliminated which, in turn, cut down the inference cost to almost half [3]. This motivated experimenting with an encoder whose number of layers is larger than the decoder's with varying extents as well as testing out the opposite to exclude any refutations. It is noteworthy to mention that we maintained the same total number of layers of the encoder and decoder combined, that is, 72 layers corresponding to 36 layers each in the previous set of experiments. Proceeding from the top-performing version so far of *Exp. 7*, we tested three variations of encoder-decoder's relative number of layers.

> **Exp. 10:** We initially experimented with having an encoder with 54 layers and a decoder with 18 layers.

> **Exp. 11:** We then experimented with having an encoder with 48 layers and a decoder with 24 layers.

> **Exp. 12:** We finally experimented with having a decoder with 48 layers and an encoder with 24 layers.

It was found that the version with a 48-layer encoder and 24-layer decoder -*Exp. 11*- outperformed the version of *Exp. 8* in all of the secondary structure prediction tasks (8-states), the fold and sub-cellular localization prediction, and the overall mean and median whilst the other two versions demonstrated a fluctuating performance depending on the task. Eventually, we proceeded with *Exp. 11* (48-layer encoder and 24-layer decoder) to the subsequent set of experiments, unlocking a gain in extracting richer embeddings -as a result of a bigger encoder- with the same total cost of equal-sized encoder-decoder. Our choice was mainly motivated by promoting generalization embodied in the need to retain an adequate number of decoder layers due to their importance in a broad class of generation tasks as well as pooling the majority of task datasets. Moreover, our choice was also motivated by computational efficiency embodied in the smaller number of encoder layers resulting in faster feature extraction. The results for this set of experiments can be found in Table 5.

**b. Depth vs. Width Variation layers** Depth corresponds to the number of layers which, as demonstrated, corresponds to the encoder and decoder's layers in the case of the transformer. Width, however, corresponds to the embedding dimension in the transformer's context [19].

> **Exp. 13:** The only experiment conducted w.r.t depth-width variations corresponds to increasing the embedding layer dimension from 768 to 1024 and accordingly reducing the depth from a 48-layered encoder & 24-layered decoder to a a 24-layered encoder & a 12-layered decoder to retain, approximately, similar or smaller number of parameters. This, however, corresponded to fluctuating results and accordingly was not proceeded with. We refer to the version with an embedding dimension of 768 as the base model or, *Ankh_base*. The results for this set of experiments can be found in Table 6.

**c. Activation Function** The Activation function is the function introducing non-linearity to the forward layer [56]. So far and throughout all the previous experiments, the activation function we have been using was Gated-GELU. It is important to note that Gated-GELU denotes a significantly larger trainable parameter size that, in turn, forces us to neutralize increases in width by decreases in depth, which we have done in *Exp. 12* [15]. To overcome the high parameter demand of the Gated-GELU forcing us to reduce the model's depth, we have ought to change the activation function altogether so that we are not mandated to compromise significantly large depths. Thus, the remaining two experiments conducted w.r.t depth-width variations correspond to a change of the activation function to classic ReLU and testing two combos of depth and width [57]. Now, it may come to mind that varying depth/width alongside the activation function corresponds to two independent variables. However, it is important to note that this is traced to the nature of the Gated-GELU proposing a demand that no longer exists when the Gated-GELU is omitted as well as the constraint of maintaining the same number of parameters to opt for a computationally-fair comparison.

**Exp. 14:** The first combo we tested is a depth of 62-layer encoder and 11-layer decoder with an embedding dimension of 768.

**Exp. 15:** The second combo is a depth of 48-layer encoder and 24-layer decoder, also with an embedding dimension of 768.

It was found that none of the combos pursued in the depth-width variation set of experiments consistently surpassed the top performer version, *Exp. 11*. Hence, none of this set's combos were proceeded with and we reverted back to *Exp. 11*. The results for this set of experiments can be found in Table 7.

**d. Relative Positional Embedding** Positional embedding describes the location of sequence tokens so that each position is assigned a unique fixed representation, an essential assignment in the case of transformers as their fundamental idea is the replacement of sequential units with attention. [58]. Using a fixed positional embedding has several limitations such as not being able to extract embeddings for tokens exceeding the maximum length of the pre-defined positional embedding dimension, correlated to the embedding dimension [59]. To overcome this limitation, Relative Positional Embedding was introduced. Instead of utilizing a fixed embedding per position, Relative Positional Embedding assigns the sequence tokens variable and relative positional representations whose variability derives from the offset between the "key" and "query" compared in the self-attention setting. We utilize a simplified variation of the Relative Positional Embedding introduced in [20] where every representation is merely a scalar added to the denoting logit utilized for computing the attention weights. In this setting, all representations' parameters are shared across all layers in our model although each attention head uses a different learned positional representation within a given layer. Although the learned embeddings/representations are relative and variable, the embedding dimension learned is fixed in the same experiment as it corresponds to a range of possible key-query offsets. So far and throughout the aforementioned experiments, we have been using 32 embeddings for all of our models. Furthermore, we have been using an offset of 128 tokens throughout which we assign all relative positions to the same embedding. It is noteworthy that a given layer is insensitive to a relative position beyond 128 tokens, but the following layers can still be sensitive to larger offsets by combining local information from previous layers, enabling the model to accommodate sequence lengths larger than the maximum predetermined length. This set of experiments corresponds to different combos of the embedding dimension as well as the embedding offset to ultimately test whether it is better to have a few large-sized embeddings or many small-sized ones, or something in between. The first two experiments in this set retained the default embedding dimension of 32 but varied the embedding offset.

**Exp. 16:** The first experiment in this set retained the default embedding dimension, that is, 32 but increased the embedding offset to 256.

**Exp. 17:** The second experiment in this set also retained the default embedding dimension of 32 but decreased the embedding offset to 64. It was found that the smaller embedding offset of 64 exceeded the performance of both 256 as well as the default value of 128. The following two experiments in this set retained the top-performing embedding offset of 64 but varied the embedding dimension.

**Exp. 18:** The third experiment in this set retained the embedding offset of 64 but increased the embedding dimension to 64.

**Exp. 19:** The fourth experiment in this set retained the embedding offset of 64 but decreased the embedding dimension to 16.

Nevertheless, none of those variations consistently exceeded the performance of the embedding dimension of 32 and embedding offset of 64. As it was shown that doubles are performing better, the final two experiments in this set tried out two combos of doubles.

**Exp. 20:** The fifth experiment tested an embedding offset of 128 and an embedding dimension of 64.

**Exp. 21:** The final experiment of this set tested an embedding offset of 256 and an embedding dimension of 128.

It was found that the combo with the most consistent and general results was that of *Exp. 20* (embedding offset of 128 and an embedding dimension of 64) and accordingly proceeded with as we refer to fold prediction when classification tasks are inconsistent and refer to *CASP12* when secondary structure dataset results are inconsistent. The results for this set of experiments can be found in Table 8.

**e. Weight Tying** Weight Tying originates from the motive of reducing the number of parameters associated with the training of the language models and accordingly, training and convergence time, and results consistency [60].

> **Exp. 22:** The mechanism by which we pursue the aforementioned motives is tying/sharing the weights and biases of the embedding and the decoder. However, it was found that this did not consistently surpass the results of the so-far top-performing model. This is traced to the difference between the input and output types due to how we masked and damask the input and output tokens, respectively, denoting higher prediction abilities from a setting with fewer parameters. The results for this set of experiments can be found in Table 9.

### 4.6.3 Dataset

In the experimentation of Prot-T5, it was found that *UniRef50* outperformed larger datasets such as *UniRef100* and *BFD* [3]. This was traced to the high quality of *UniRef50* in means of the lack of duplication and sequence diversity. Throughout all the previous experiments, the pre-training was also conducted on UniRef50. Yet, we wanted to test a dataset of intermediate size between UniRef50 and the bigger datasets which have proven less efficient.

> **Exp. 23:** We have conducted a single experiment on *UniRef90*. Nonetheless, the experiment has encouraged the initial direction of proceeding with *UniRef50* as representative for efficient high-quality attributes. It is noteworthy to mention that all previous experiments with *UniRef50* were trained for 2 full epochs in contrast to the experiment with *UniRef90* which was trained for only one epoch that is arguably equivalent. The results for this set of experiments can be found in Table 10.

## 5 Availability

Both the Ankh and *Ankh_base* models are publicly available for research innovation at our Ankh repository "https://github.com/agemagician/Ankh", under an Attribution- NonCommercial-ShareAlike 4.0 International (CC BY-NC-SA 4.0) license. The repository also contains extensive python and Jupyter python notebook tutorials for various examples, including embedding extractions and supervised learning of several downstream tasks using freely available online resources (GoogleColab). For commercial usage and licensing, please visit "https://www.proteinea.com/"

## 6 Acknowledgments

The authors would like to thank the deep learning and bioinformatics teams in Proteinea for their invaluable help with hardware, software, and with many other aspects of this work. Mohammed AlQuraishi, Columbia University, for their feedback. From Google, the authors would like to thank Jonathan Caton, Shira Genauer, Astitva Chopra, and all Google cloud, Google innovator, JAX, and TRC Teams for helping to set up the project on Google Cloud and solving Google cloud issues. The models trained in this work could not be easily publicly available without support from the HuggingFace team, including Patrick von Platen, Julien Chaumond, and Clement Delangue. Google supported this project through Google research innovator and Google TPU Cloud Research Credits Program. We would like to thank all researchers worldwide who made all the datasets used in this research publicly available. Finally, ElNaggar and Essam would like to thank Allah for giving the strength, knowledge, and courage to finish this project and share it with the rest of humanity.

# 7 Supplementary Materials

## 7.1 Pre-Training Data & Model Experimentation

In this subsection, we report the statistics of the pre-training data as well as the results of every class of experimentation. In Table 2, we show the statistics of the pre-training data.

**Table 2**: Pre-training Data Statistics.

| Pre-training Data | UniRef50 | UniRef90 | UniRef100 |
|---|---|---|---|
| **Number of Proteins [in m]** | 45 | 144 | 216 |
| **Number of Residues [in b]** | 14 | 49 | 88 |
| **Disk Space [in GB]** | 16 | 54 | 150 |

Pre-training Datasets Statistics. UniRef50 and UniRef90 cluster the UniRef100 database at sequence identity thresholds of 50% and 90%, respectively. The statistics describe the number of proteins in millions (m), the number of amino acids in billions (b), and disk space in GB (uncompressed storage as text).

In Table 3 we report the results of the experimentation on the masking strategy. It can be deduced that the top performing version of the six tested versions was the version of *Exp. 4*, where we reconstruct all the consecutive unmasked tokens as a single merged token. Therefore, this was the version that carried on to the following sub-set of experiments, masking probability.

**Table 3**: Masking Strategy Experiment Statistics

| Exp. | SSP (Q3) | | | SSP (Q8) | | | CP(L) | | SLP | FolP | Avg. | Med. |
|---|---|---|---|---|---|---|---|---|---|---|---|---|
| | *CASP12* | *TS115* | *CB513* | *CASP12* | *TS115* | *CB513* | *L1* | *L5* | | | | |
| **Baseline** | 74.6% | 80.6% | 78.6% | 62.0% | 68.9% | 64.6% | 18.8% | 29.9% | 75.8% | 41.4% | 64.0% | 68.9% |
| **Exp. 1** | 75.1% | 81.6% | 80.4% | 61.3% | 69.9% | 66.0% | 20.2% | 33.6% | 74.9% | 42.6% | 65.0% | 69.9% |
| **Exp. 2** | 71.9% | 75.1% | 73.1% | 57.5% | 62.8% | 58.5% | 11.4% | 19.9% | 68.7% | 35.5% | 58.1% | 62.8% |
| **Exp. 3** | 69.5% | 74.9% | 73.5% | 54.9% | 61.3% | 57.4% | 9.3% | 14.8% | 70.4% | 28.5% | 56.1% | 61.3% |
| **Exp. 4** | **76.3%** | **82.8%** | **82.2%** | **63.7%** | **71.8%** | **68.8%** | **22.2%** | **37.6%** | **77.0%** | **50.7%** | **67.9%** | **71.8%** |
| **Exp. 5** | 75.6% | 81.0% | 80.0% | 62.2% | 68.9% | 65.8% | 18.4% | 31.5% | 73.5% | 33.4% | 63.5% | 68.9% |
| **Exp. 6** | 74.9% | 81.6% | 79.3% | 61.5% | 69.3% | 64.8% | 20.7% | 34.9% | 75.5% | 46.9% | 65.4% | 69.3% |

Results for the masking strategy set of experiments.

In Table 4, we report the results of the experimentation on the masking probability. Out of the four values, 10% was the worst masking probability and was accordingly disregarded. Interestingly, it was found that the default value of Exp. 4 (15% masking probability) was outperforming for the sub-cellular localization prediction, fold prediction, as well as some of the secondary structure prediction tasks for datasets such as CB513 and TS115. Nevertheless, it was found that the value of Exp. 9 (30% masking probability) was outperforming for the entirety of the contact prediction tasks in addition to the secondary structure prediction tasks for CASP12 dataset. Given the inconsistency amongst the secondary structure prediction dataset results, we referred to the results on CASP12 -being a domain standard- that promotes the higher masking probability. Furthermore and as we are opting for holistic predictions via a general-purpose language model, we promoted generalization across different types of tasks, especially when the difference amongst results is of such a small magnitude and when acknowledging that task- specific customization is of impact. Finally, given that all the experimental variations were trained for only two epochs, we proceeded with the intermediate value of Exp. 8 (20% masking probability) for the post-experimentation long-term training to fulfill the semi-comprehensive inclusion of different polls of tasks anticipated from a general-purpose protein language model that can then be customized per downstream task.

**Table 4**: Masking Probability Experiment Statistics

| Exp. | SSP (Q3) | | | SSP (Q8) | | | CP(L) | | SLP | FolP | Avg. | Med. |
|---|---|---|---|---|---|---|---|---|---|---|---|---|
| | *CASP12* | *TS115* | *CB513* | *CASP12* | *TS115* | *CB513* | *L1* | *L5* | | | | |
| **Exp. 4** | 76.3% | 82.8% | **82.2%** | 63.7% | **71.8%** | **68.8%** | 22.2% | 37.6% | **77.0%** | **50.7%** | 67.9% | **71.8%** |
| **Exp. 7** | 76.7% | **83.0%** | 81.9% | 63.7% | 71.7% | 68.5% | 21.8% | 33.8% | 76.7% | 46.9% | 66.7% | 71.5% |
| **Exp. 8** | 75.9% | 82.7% | 81.4% | 62.8% | 71.5% | 67.9% | 22.3% | 35.1% | 76.7% | 47.6% | 67.1% | 71.7% |
| **Exp. 9** | **76.8%** | 82.2% | 81.3% | **64.0%** | 71.4% | 68.0% | **24.1%** | **37.9%** | 75.1% | 46.3% | 67.0% | 71.4% |

Results for the masking probability set of experiments.

In Table 5, we report the results of the experimentation on the number of encoder and decoder layers. It was found that the version with a 48-layer encoder and 24-layer decoder -*Exp. 11*- outperformed the version of *Exp. 8* in all of the secondary structure prediction tasks (8-states), the fold and sub-cellular localization prediction, and the overall mean and median whilst the other two versions demonstrated a fluctuating performance depending on the task. Eventually, we proceeded with *Exp. 11* (48-layer encoder and 24-layer decoder) to the subsequent set of experiments, unlocking a gain in extracting richer embeddings -as a result of a bigger encoder- with the same total cost of equal-sized encoder-decoder. Our choice was mainly motivated by promoting generalization embodied in the need to retain an adequate number of decoder layers due to their importance in a broad class of generation tasks as well as pooling the majority of task datasets. Moreover, our choice was also motivated by computational efficiency

**Table 5**: Architecture Experiments: No. of Encoder and Decoder Layers

| Exp. | SSP (Q3) | | | SSP (Q8) | | | CP(L) | | SLP | FolP | Avg. | Med. |
|---|---|---|---|---|---|---|---|---|---|---|---|---|
| | *CASP12* | *TS115* | *CB513* | *CASP12* | *TS115* | *CB513* | *L1* | *L5* | | | | |
| Exp. 8 | 75.9% | 82.7% | 81.4% | 62.8% | 71.5% | 67.9% | 22.3% | 35.1% | 76.7% | 47.6% | 67.1% | 71.7% |
| Exp. 10 | 76.5% | **83.2%** | 82.0% | 63.2% | 72.2% | 68.6% | **24.0%** | **39.5%** | 76.8% | 48.1% | 67.7% | 72.2% |
| Exp. 11 | 76.4% | 83.1% | **82.3%** | **63.7%** | **72.3%** | **69.2%** | 23.0% | 37.1% | **77.2%** | **48.5%** | **67.8%** | **72.3%** |
| Exp. 12 | **76.8%** | 82.9% | 81.4% | 63.3% | 71.8% | 68.1% | 21.3% | 34.9% | 75.7% | 46.5% | 66.8% | 71.8% |

Results for the number of encoder and decoder layers' set of experiments.

embodied in the smaller number of encoder layers resulting in faster feature extraction.

In Table 6, we report the results of the experimentation on the depth vs. width variations. We can see that the tested variation corresponded to fluctuating results and accordingly was not proceeded with. We refer to the version with an embedding dimension of 768 as the base model or, *Ankh_base*.

**Table 6**: Architecture Experiments: Depth vs. Width Variations

| Exp. | SSP (Q3) | | | SSP (Q8) | | | CP(L) | | SLP | FolP | Avg. | Med. |
|---|---|---|---|---|---|---|---|---|---|---|---|---|
| | *CASP12* | *TS115* | *CB513* | *CASP12* | *TS115* | *CB513* | *L1* | *L5* | | | | |
| Exp. 11 | **76.4%** | 83.1% | **82.3%** | **63.7%** | **72.3%** | **69.2%** | **23.0%** | 37.1% | 77.2% | **48.5%** | **67.8%** | **72.3%** |
| Exp. 13 | 76.1% | **83.3%** | 82.6% | 62.7% | 70.9% | 68.0% | 22.6% | **37.5%** | **77.3%** | 47.8% | 67.4% | 70.9% |

Results for the depth vs. width variation experiments.

In Table 7, we report the results of the experimentation on the activation function.

**Table 7**: Architecture Experiments: Activation Function

| Exp. | SSP (Q3) | | | SSP (Q8) | | | CP(L) | | SLP | FolP | Avg. | Med. |
|---|---|---|---|---|---|---|---|---|---|---|---|---|
| | *CASP12* | *TS115* | *CB513* | *CASP12* | *TS115* | *CB513* | *L1* | *L5* | | | | |
| Exp. 11 | **76.4%** | **83.1%** | **82.3%** | **63.7%** | **72.3%** | **69.2%** | **23.0%** | **37.1%** | **77.2%** | **48.5%** | **67.8%** | **72.3%** |
| Exp. 14 | **76.4%** | **83.1%** | 78.9% | 62.7% | 69.0% | 64.7% | 19.7% | 33.4% | 76.0% | 45.4% | 65.3% | 69.0% |
| Exp. 15 | 75.8% | 82.0% | 80.1% | 62.5% | 70.8% | 66.6% | 21.2% | 36.7% | 75.5% | 46.6% | 66.3% | 70.8% |

Results for the activation function experiments.

It was found that none of the combos pursued in the depth-width variation set of experiments consistently surpassed the top performer version, *Exp. 11*. Hence, none of this set's combos were proceeded with and we reverted back to *Exp. 11*.

In Table 8, we report the results of the experimentation on the relative positional encoding. It was found that the combo with the most consistent and general results was that of *Exp. 20* (embedding offset of 128 and a number of embeddings of 64) and accordingly proceeded with as we refer to fold prediction when classification tasks are inconsistent and refer to *CASP12* when secondary structure dataset results are inconsistent.

In Table 9, we report the results of the experimentation on the weight tying. It can be seen that this experiment did not consistently surpass the results of the so-far top-performing model. This is traced to the difference between the input and output types due to how we masked and damask the input and output tokens, respectively, denoting higher prediction abilities from a setting with fewer parameters.

**Table 8**: Architecture Experiments: Relative Positional Encoding

| Exp. | SSP (Q3) | | | SSP (Q8) | | | CP(L) | | SLP | FoIP | Avg. | Med. |
|---|---|---|---|---|---|---|---|---|---|---|---|---|
| | *CASP12* | *TS115* | *CB513* | *CASP12* | *TS115* | *CB513* | *L1* | *L5* | | | | |
| **Exp. 15** | 76.4% | 83.1% | **82.3%** | 63.7% | **72.3%** | **69.2%** | 23.0% | 37.1% | 77.2% | 48.5% | 67.8% | **72.3%** |
| **Exp. 16** | 76.2% | 82.9% | 82.0% | 64.0% | 71.8% | 68.9% | 22.7% | 38.0% | **77.8%** | 47.6% | 67.7% | 71.8% |
| **Exp. 17** | 76.7% | 83.0% | 82.0% | 63.7% | 71.7% | 68.8% | 23.6% | **40.8%** | 76.8% | 50.1% | 68.2% | 71.7% |
| **Exp. 18** | 76.1% | **83.3%** | 82.0% | 63.7% | 71.7% | 68.6% | 21.5% | 34.9% | 76.9% | 48.4% | 63.7% | 71.7% |
| **Exp. 19** | 76.0% | **83.3%** | 81.9% | 62.9% | 72.1% | 68.6% | **24.2%** | 40.2% | 77.2% | 48.0% | 67.8% | 72.1% |
| **Exp. 20** | **77.0%** | 83.0% | **82.3%** | **64.3%** | 72.2% | 69.1% | 23.6% | 40.7% | 77.2% | **51.8%** | **68.6%** | **72.2%** |
| **Exp. 21** | 76.2% | 82.5% | 81.5% | 63.9% | 71.1% | 68.1% | 25.2% | 41.3% | 76.4% | 49.2% | 67.8% | 71.1% |

Results for the relative positional encoding experiments.

**Table 9**: Architecture Experiments: Weight Tying

| Exp. | SSP (Q3) | | | SSP (Q8) | | | CP(L) | | SLP | FoIP | Avg. | Med. |
|---|---|---|---|---|---|---|---|---|---|---|---|---|
| | *CASP12* | *TS115* | *CB513* | *CASP12* | *TS115* | *CB513* | *L1* | *L5* | | | | |
| **Exp. 20** | **77.0%** | **83.0%** | **82.3%** | **64.3%** | **72.2%** | **69.1%** | **23.6%** | **40.7%** | 77.2% | **51.8%** | **68.6%** | **72.2%** |
| **Exp. 22** | 75.9% | **83.0%** | **82.3%** | 63.0% | 72.1% | **69.1%** | 22.6% | 36.0% | **78.1%** | 48.9% | 67.6% | 72.1% |

Results for the weight tying experiments.

In Table 10, we report the results of the experimentation on the dataset. Nonetheless, the experiment has encouraged the initial direction of proceeding with *UniRef50* as representative for efficient high-quality attributes. It is noteworthy to mention that all previous experiments with *UniRef50* were trained for 2 full epochs in contrast to the experiment with *UniRef90* which was trained for only one epoch that is arguably equivalent.

**Table 10**: Dataset Experiments

| Exp. | SSP (Q3) | | | SSP (Q8) | | | CP(L) | | SLP | FoIP | Avg. | Med. |
|---|---|---|---|---|---|---|---|---|---|---|---|---|
| | *CASP12* | *TS115* | *CB513* | *CASP12* | *TS115* | *CB513* | *L1* | *L5* | | | | |
| **Exp. 20** | **77.0%** | **83.0%** | **82.3%** | **64.3%** | **72.2%** | 69.1% | **23.6%** | **40.7%** | **77.2%** | **51.8%** | **68.6%** | **72.2%** |
| **Exp. 23** | 75.9% | 82.8% | 81.9% | 63.3% | 71.1% | **69.4%** | 22.8% | 38.7% | 75.4% | 45.9% | 67.2% | 71.7% |

Results for the dataset experiments.

## 7.2  Model Pre-training

In this subsection, we report the configurations and hyper-parameters associated with the pre-training of the two Ankh models in comparison to the baseline as well as the downstream model in Table 11.

**Table 11**: Pre-training Configurations.

| Hyper-parameter | Baseline | Ankh_base | Ankh_large | Downstream Models |
|---|---|---|---|---|
| **Dataset** | UniRef50 | UniRef50 | UniRef50 | NA |
| **Embedding Dimension** | 768 | 768 | 1536 | Ankh Emb. Dim. |
| **No. of Attention Heads** | 12 | 12 | 16 | 4 |
| **Learning Rate** | 0.01 | 0.004 | 0.002 | 0.001 |
| **No. of Training Epochs** | 2 | 68 | 68 | 5 |
| **Masking Probability** | 15% | 20% | 20% | NA |
| **No. of Encoder Layers** | 36 | 48 | 48 | NA |
| **No. of Decoder Layers** | 36 | 24 | 24 | NA |
| **Activation Function** | Gated-GELU | Gated-GELU | Gated-GELU | Gated-GELU |
| **Maximum length** | 512 | 512 | 512 | NA |
| **Local Batch Size** | 16 | 24 | 24 | 1 |
| **Global Batch Size** | 960 | 1536 | 1024 | 16 |
| **Optimizer** | Adafactor | Adafactor | Adafactor | AdamW |
| **Scheduler** | Linear | Linear | Linear | Linear |
| **No. of Feed-forward Layers** | 3072 | 3072 | 3840 | Emb. Dim./2 |
| **Warm-Up Steps** | 10k | 10k | 10k | 1k |
| **Weight Decay** | None | None | None | None |
| **Random Seed** | 42 | 42 | 42 | 7 |
| **TPUs/GPUs** | V4-64 | V4-128 | V4-128 | A100 |
| **Weight/Bias Initialization** | random | random | random | random |

The configurations used to the experimentation baseline model, pre-trained *Ankh_base*, *Ankh*, and top model used in the downstream tasks.

## 7.3 Data & Model Testing

### 7.3.1 Downstream Dataset

In this subsection, we report the statistics of the multiple datasets used in the downstream tasks. The statistics can be observed in Table 12.

**Table 12**: Description of Supervised Downstream Datasets.

| Task Name | Task Category | Output Distribution | Modeling Level | Data Source | # Proteins | Split Source | # Train/Valid./Test |
|---|---|---|---|---|---|---|---|
| FluP | Function | Regression | Protein-level | Sarkisyan's | 54,025 | TAPE | 21,446/5,362/27,217 |
| SolP | Function | Classification | Protein-level | DeepSol | 71,419 | DeepSol | 62,478/6,942/1,999 |
| GB1P | Function | Regression | Protein-level | FLIP | 8,733 | FLIP | 6,291/670/1,772 |
| CP | Structure | Classification | Residue Pair level | ProteinNet | 25,563 | TAPE | 25,299/224/40 |
| FolP | Structure | Classification | Protein-level | DeepSF | 13,766 | DeepSF | 12,312/736/718 |
| SSP | Structure | Classification | Residue-level | NetSurfP-2.0 | 11,361 | TAPE | 8,678/2,170/513 |
| LocP | Localization | Classification | Protein-level | DeepLoc | 13,961 | DeepSF | 8,945/2,248/2,768 |

A description of Benchmark tasks investigated in the comparative study. We report categorical information about the tasks, alongside details about the original source of the task datasets, number of proteins, and the source and numerical distribution of the training/validation/testing splits utilized. Abbr., FluP: Fluorescence Prediction; StaP: Thermal Stability Prediction; SolP: Solubility Prediction; CP: Contact Prediction; FolP: Fold Prediction; SSP: Secondary Structure Prediction; LocP: Localization Prediction.

### 7.3.2 Full Results of Embedding-Based Predictions

Besides the domain standard datasets presented in our main pages, this section shows in Table 13 the full results of the embedding-based predictions spanning a variety of additional datasets and also confirming the same results.

**Table 13**: Results Summary.

| Task | Dataset | Ankh | Ankh_base | ProtT5-XL-U50 | ESM-1b | ESM-2 (650M) | ESM-2 (3B) | ESM-2 (15B) |
|---|---|---|---|---|---|---|---|---|
| SSP (Q3) | CASP12 | **83.8**±3% | 80.8±4% | 83.4±4% | 79.6±4% | 82.3±4% | 83.3±4% | 83.2±3% |
| | CASP14 | **77.6**±3% | 76.8±3% | 74.1±3% | 75.1±4% | 77.0±3% | 76.8±3% | 76.8±4% |
| | TS115 | **88.2**±1% | 86.8±1% | 86.8±1% | 84.9±1% | 87.2±1% | 87.5±1% | 87.3±1% |
| | CB513 | **88.6**±0.6% | 87.0±0.7% | 86.7±0.6% | 84.4±0.7% | 87.0±0.6% | 87.3±0.6% | 87.4±0.6% |
| SSP (Q8) | CASP12 | **71.5**±4% | 68.7±4% | 70.7±3% | 66.6±4% | 70.6±3% | 71.4±4% | 71.4±4% |
| | CASP14 | **63.2**±3% | 62.3±4% | 60.7±5% | 60.0±5% | 62.7±4% | 61.6±5% | 61.7±6% |
| | TS115 | **78.2**±2% | 76.9±2% | 77.0±2% | 73.7±2% | 77.0±2% | 77.2±2% | 77.6±2% |
| | CB513 | **77.4**±1% | 75.5±1% | 75.1±1% | 71.0±1% | 75.6±1% | 75.8±1% | 75.9±1% |
| CP (L) | ProteinNet L/1 | **49.0**±8% | 43.2±8% | 44.7±8% | 30.0±6% | 29.6±6% | 30.7±6% | 33.3±6% |
| | ProteinNet L/5 | **73.2**±11% | 66.6±11% | 69.2±11% | 50.1±10% | 50.2±10% | 52.7±10% | 54.7±10% |
| | CASP14 L/1 | **30.2**±8% | 28.8±7% | 26.9±7% | 24.6±6% | 25.0±6% | 24.8±7% | 25.9±7% |
| | CASP14 L/5 | **50.7**±11% | 48.0±11% | 42.4±14% | 40.0±11% | 38.4±13% | 41.9±14% | 40.4±15% |
| EAT | | 71.7±6% | **74.8**±6% | 71.0±6% | 64.5±7% | 55.5±7% | 65.6±6% | 65.4±7% |
| FolP | | **61.1**±4% | 58.8±4% | 57.6±4% | 57.6±4% | 56.3±4% | 60.5±4% | 56.7±4% |
| GB1P | | 0.84±0.008 | **0.85**±0.008 | 0.78±0.01 | 0.81±0.009 | 0.82±0.009 | 0.81±0.009 | 0.57±0.02 |
| LocP | | **83.2**±2% | 81.4±2% | **83.2**±2% | 80.0±2% | 81.8±2% | 82.4±2% | 81.8±2% |
| FluP | | **0.62**±0.4 | 0.61±0.4 | 0.58±0.4 | 0.5±0.5 | 0.48±0.5 | 0.48±0.5 | 0.55±0.4 |
| SolP | | **76.4**±2% | 74.2±2% | 74.4±2% | 67.3±2% | 75.0±2% | 74.9±2% | 74.4±2% |
| Avg. | | **71.43%** | 70.0% | 69.3% | 65.3% | 66.9% | 67.7% | 66.6% |
| Med. | | **77.5%** | 76.0% | 75.2% | 71.9% | 76.1% | 76.4% | 73.5% |

Summary of the benchmarking results involving, *Ankh* and *Ankh_base*, with ProtT5-XL-U50, ESM-1b, ESM-2 (650M), ESM-2 (3B), ESM-2 (15B). We report the Spearman Correlation score for the regression tasks and accuracy scores for all classification tasks, except for contact prediction where we report L/1 and L/5 precision. In EAT, the score reported is the mean of the accuracy scores of the four annotations (Class, Architecture, Topology and Homologous superfamily). Task Abbreviations: SSP: Secondary Structure Prediction; CP: Contact Prediction; EAT: Embedding-based Annotation Transfer; FolP: Fold Prediction; GB1P: GB1 Fitness Prediction; LocP: Localization Prediction; FluP: Fluorescence Prediction; SolP: Solubility Prediction.

### 7.3.3 Attention-Based Predictions

In this subsection, we report the results of the attention-based prediction done on the contact prediction task in Table 14. As attention maps are reported as the better indicator of contact prediction for the ESM PLM family in [2] and [10], the two representations are tested. To elaborate, the attention maps are extracted and compared with the contextualized embeddings as input for the contact prediction task per every model to opt for fair comparison and demonstrate the SoA's best indicator with what we deem as the best indicator.

**Table 14**: Attention-Based Predictions.

| Task | Dataset | Ankh_large | Ankh_base | ProtT5-XL-U50 | ESM-1b | ESM-2 (650M) | ESM-2 (3B) | ESM-2 (15B) |
|---|---|---|---|---|---|---|---|---|
| **CP (L)** | *ProteinNet L/1* | **31.4**±5% | 25.9±5% | 30.9±5% | 25.3±5% | 31.9±6% | 33.9±6% | 33.3±6% |
| | *ProteinNet L/5* | **55.6**±8% | 46.3±8% | 51.9±8% | 42.0±8% | 54.6±9% | 56.6±9% | 57.4±9% |
| | *CASP14 L/1* | **11.1**±6% | 9.3±5% | 8.6±5% | 7.8±4% | 10.7±5% | 12.2±7% | 12.3±7% |
| | *CASP14 L/5* | **20.7**±12% | 19.5±10% | 16.1±10% | 15.8±10% | 21.0±11% | 21.4±13% | 24.6±13% |

The results for benchmarking the attention-based predictions, which showed worse performance than the embedding-based contact prediction results

### 7.3.4 Embedding-Based Annotation Transfer

In this subsection, we report the full results of the embedding-based annotation transfer for the CATH domains as seen in Table 15.

**Table 15**: The accuracy scores of the embedding-based attention transfer

| Model | C | A | T | H | Mean |
|---|---|---|---|---|---|
| ProtT5-XL-U50 | 83.6±5% | 70.0±6% | 57.2±7% | **73.3**±7% | 71.0±6% |
| ESM-1b | 78.1±5% | 65.3±6% | 51.4±7% | 63.3±8% | 64.5±7% |
| ESM-2 (650M) | 72.1±6% | 56.2±7% | 40.4±6% | 53.3±8% | 56.0±7% |
| ESM-2 (3B) | 79.5±5% | 65.8±6% | 53.4±7% | 64.0±8% | 65.6±6% |
| ESM-2 (15B) | 78.5±5% | 63.5±6% | 53.4±7% | 67.3±8% | 65.4±7% |
| *Ankh_base* | **85.8**±5% | **77.2**±6% | **63.5**±6% | 72.7±7% | **74.8**±6% |
| *Ankh_large* | 83.6±5% | 72.1±6% | 61.0±7% | 70.7±7% | 71.7±6% |

The accuracy scores reported over the four CATH levels comparing the truth labels with the predicted labels corresponding to the closest match in the lookup embeddings space. Level 1 C is for Class. Level 2 A is for Architecture. Level 3 T is for Topology. Level 4 H is for Homologous Superfamily.

### 7.3.5 CATH Domains

In this subsection, we report the results of the CATH domain analysis on the natural and generated data. In Table 16, we observe the CATH super-domain analysis.

**Table 16**: A summary of the CATH Super-Family Domains in Generated vs. Natural (Fine-Tuning) Data.

| Name | Super-family ID | Natural | t1.e1 | t1.e2 | t2.e1 | t2.e2 |
|---|---|---|---|---|---|---|
| Lactate dehydrogenase/glycoside hydrolase | 3.90.110.10 | 8088 | 4549 | 3961 | 6103 | 6657 |
| AglA-like glucosidase | 3.90.1820.10 | 7540 | 4082 | 3648 | 5337 | 5863 |
| NAD(P)-binding Rossmann-like Domain | 3.40.50.720 | 171 | 24 | 30 | 3059 | 2148 |
| Other Domains | NA | 196 | 0 | 0 | 18 | 5 |
| Total Domains | NA | 15799 | 8655 | 7639 | 14517 | 14673 |

The distribution of CATH domains belonging to the three dominating homologous superfamilies in the natural set. Abbr., Nat: Natural Sequences; t1.e1: Generated Sequences at temperature 1 by the first epoch; t1.e2: Generated Sequences at temperature 1 by the second epoch; t2.e1: Generated Sequences at temperature 2 by the first epoch; t2.e2: Generated Sequences at temperature 2 by the second epoch.

In Table 17, we observe the CATH functional-domain analysis.

**Table 17**: A summary of the CATH Functional-Family Domains in Generated vs. Natural (Fine-Tuning) Data.

| Name | Functional-family No. | Super-family ID | Natural | t1.e1 | t1.e2 | t2.e1 | t2.e2 |
|---|---|---|---|---|---|---|---|
| | 4 | 3.90.110.10 | 3675 | 2081 | 1917 | 3150 | 3277 |
| | 18 | 3.40.50.720 | 170 | 23 | 5 | 1772 | 2647 |
| **Malate dehydrogenase** | 1 | 3.90.110.10 | 6 | 0 | 0 | 155 | 95 |
| | 17 | 3.40.50.720 | 0 | 0 | 0 | 8 | 6 |
| **Sum of Malate dehydrogenase** | NA | NA | 6812 | 3582 | 3621 | 6553 | 7931 |
| | 11 | 3.90.110.10 | 421 | 205 | 125 | 282 | 254 |
| | 3 | 3.90.110.10 | 28 | 89 | 30 | 77 | 43 |
| **L-lactate dehydrogenase** | 29 | 3.40.50.720 | 0 | 0 | 0 | 13 | 20 |
| | 362 | 3.40.50.720 | 0 | 0 | 0 | 0 | 1 |
| **Sum of L-lactate dehydrogenase** | NA | NA | 449 | 294 | 155 | 372 | 318 |

The distribution of the domains with known functional annotations distributed over the different Fun-Family numbers. The majority of the natural domains belong are annotated as malate dehydrogenase (MDH), the protein family of interest. A smaller portion of the domains belongs to a closely similar protein family L-lactate dehydrogenase (LDH), which is close functionality to MDH but it is not the optimization goal. The generated sets at high tempratures are biased towards generating MDH domains compared with LDH domains.